\documentclass[]{TEAI}
\usepackage{helvet}

\usepackage{amsmath} 
\usepackage{natbib}
\usepackage{graphicx}
\usepackage{subcaption} 

\usepackage[toc,page,header]{appendix}
\usepackage[utf8]{inputenc} 
\usepackage[T1]{fontenc}    
\usepackage{url}            
\usepackage{booktabs}       
\usepackage{lmodern}        
\usepackage{amsfonts}       
\usepackage{nicefrac}       
\usepackage{microtype}      
\usepackage{wrapfig}

\usepackage{amssymb}  
\usepackage{fontawesome}  
\usepackage{url}  

\usepackage{titletoc}

\usepackage{tikz}  
\usepackage{comment}  
\usepackage{tabularx}  
\usepackage{booktabs}  

\usepackage{minitoc}

\usepackage{booktabs}
\usepackage{array}
\usepackage{etoolbox}

\definecolor{lightblue}{RGB}{200, 230, 255}  
\definecolor{headerblue}{RGB}{150, 200, 255} 

\usepackage{pgfplots}
\usepackage[utf8]{inputenc} 
\usepackage[T1]{fontenc}    
\usepackage{url}            
\usepackage{booktabs}       
\usepackage{amsfonts}       
\usepackage{nicefrac}       
\usepackage{microtype}      
\usepackage{xcolor}         
\usepackage{graphicx}
\usepackage{float}
\usepackage{comment}
\usepackage{multirow} 
\usepackage{amsmath} 
\usepackage{makecell} 
\usepackage{siunitx}  
\usepackage{tikz}
\usepackage{pgf-pie} 
\usepackage{subcaption}
\usepackage{wrapfig}
\usepackage[export]{adjustbox}

\usepackage{ragged2e}      
\usepackage{tabularx}       
\usepackage{array}          
\usepackage{caption}        
\usepackage{enumitem}
\usepackage{pifont}
\usepackage[hang,flushmargin]{footmisc} 

\usepackage{tcolorbox}

\usepackage{tcolorbox}    
\tcbuselibrary{breakable}  
\tcbuselibrary{skins}      

\usepackage{tabularx}
\usepackage{listings}

\usepackage[utf8]{inputenc} 
\usepackage[T1]{fontenc}    
\usepackage{url}            
\usepackage{booktabs}       
\usepackage{amsfonts}       
\usepackage{nicefrac}       
\usepackage{microtype}      
\usepackage[table]{xcolor}         
\usepackage{todonotes}
\usepackage{multirow}   
\usepackage{makecell}   
\usepackage{graphicx}   
\usepackage{caption} 
\usepackage{amsmath}
\usepackage{subcaption}  
\usepackage{placeins}
\usepackage{acronym}
\usepackage{enumitem}
\usepackage{float}
\usepackage{algorithm}
\usepackage{algpseudocode}
\usepackage{wrapfig}
\usepackage{placeins}

\DeclareSymbolFont{cmexlargesymbols}{OMX}{cmex}{m}{n}
\DeclareMathSymbol{\sumop}{\mathop}{cmexlargesymbols}{"50}

\definecolor{nipscolor}{rgb}{0.21,0.49,0.74}
\hypersetup{
    colorlinks=true,
    linkcolor=red,
    citecolor=nipscolor
}

\usepackage[most]{tcolorbox}
\usepackage{fvextra}

\newcolumntype{C}[1]{>{\centering\arraybackslash}p{#1}}

\title{VLA-Pro: Cross-Task Procedural Memory Transfer for Vision-Language-Action Models}

\renewcommand{\authorlist}{%
    {\authorfont
    \textbf{Shengyu Si}$^{1,2*}$\quad
    \textbf{Yuanzhuo Lu}$^{1,2*}$\quad
    \textbf{Ruimeng Yang}$^{1,2*}$\\[0.3em]
    \textbf{Ziyi Ye}$^{1,2\dagger}$\quad
    \textbf{Zuxuan Wu}$^{1,2,3}$\quad
    \textbf{Yu-Gang Jiang}$^{1,2}$}%
}

\renewcommand{\affiliationlist}{%
    {\affiliationfont
    $^{1}$\mbox{Institute of Trustworthy Embodied AI, Fudan University}\\[0.3em]
    $^{2}$\mbox{Shanghai Key Laboratory of Multimodal Embodied AI}
    \quad
    $^{3}$\mbox{Neote AI}\\[0.3em]
    \texttt{sysi25@m.fudan.edu.cn}\quad
    \texttt{zyye@fudan.edu.cn}}%
}

\providecommand{\abstractlist}{}
\renewcommand{\abstractlist}{%
{\abstractinfont
Vision-Language-Action~(VLA) models have shown strong potential for general-purpose robotic manipulation, yet they still struggle to generalize to unseen tasks that necessitate transferring relevant experience across objects, scenes, and action patterns.
This paper proposes VLA-Pro, a plug-and-play framework designed to enhance cross-task generalization by storing task-relevant procedural memories at training time and transferring these memories during inference.
Specifically, VLA-Pro represents each procedural memory as a structured procedural-state index paired with a task-specific LoRA adapter.
At inference time, VLA-Pro retrieves relevant procedural memories based on the current multi-modal context and dynamically fuses these memories for generating the current action chunk.
Experiments on RoboTwin, RLBench, and real-world manipulation tasks show that VLA-Pro consistently improves cross-task generalization across multiple backbones, achieving up to a 207\% relative improvement in simulation and increasing real-world success rate from 5.8\% to 65.0\%.
These results suggest that procedural memory retrieval and adaptation provide an effective mechanism for transferring manipulation experience to novel tasks while preserving modularity and execution stability.
}
}

\renewcommand{\checkdatalist}{%
    {\small
    {\checkdatafont\bfseries Code:} \url{https://github.com/ketchup45/VLA-Pro}\par
    {\checkdatafont\bfseries Website:} \url{https://ketchup45.github.io/VLA-Pro}\par}%
}

\begin{document}
\maketitle
\renewcommand{\thefootnote}{}
\footnotetext{$^*$Equal Contribution.\quad
              $^\dagger$Corresponding author.}
\renewcommand{\thefootnote}{\arabic{footnote}}


\vspace{-1.5em}

\section{Introduction}

Current vision-language-action (VLA) models~\cite{black2024pi_0,black2025pi_,brohan2023rt,driess2023palm,team2024octo,zitkovichrt} have made remarkable progress in a wide range of daily and industrial tasks~\cite{goyal2024rvt}.
By adding an additional action head to a vision-language model, VLAs can follow natural language instructions and perform non-programmed tasks in an end-to-end manner\cite{yangvision}. 
Despite these advances, existing VLAs still struggle severely with cross-task generalization: they often fail to adapt to unseen tasks that involve novel objects and unfamiliar scenes~\cite{song2026reconvla,ye2026st4vla}, even when such tasks are semantically or physically similar to those encountered during training.

In contrast, humans exhibit far stronger generalization capabilities to tackle novel situations.
This is partially enabled by a specialized mechanism known as procedural memory ~\cite{dai2026robomme,ju2024robo,sidik20263d,zhang2026recurrent}, which allows humans to implicitly leverage past experience.
For example, when grasping a lying bottle, humans may intuitively recall the experience of holding cylindrical lying objects such as microphones and folded umbrellas, using that familiar physical ``feel'' to control grip force and orientation~\cite{ze20243d}. 
More importantly, recalling task-specific memories can counteract execution biases toward high-frequency actions~(e.g., preventing the model from defaulting to ``grasping an upright bottle'' when it specifically requires manipulating a lying one). 

Motivated by humans' memory mechanism~\cite {christie2026local,kuangram,xie2026zero}, recent efforts that improve VLA's adaptability can be broadly categorized into two primary paradigms: architecture restructuring and contextual prompting.
Architecture-based approaches encode reusable execution priors through pre-defined architectural modules~\cite{cen2025rynnvla,mei2025omnirouter}.
For instance, AtomicVLA~\cite{zhang2026atomicvla} employs skill-guided MoE to organize atomic skill experts, while MemoryVLA~\cite{shi2025memoryvla} and Chameleon~\cite{guo2026chameleon} introduce perceptual-cognitive mechanisms to handle temporal dependencies.
On the other hand, prompt-based approaches leverage demonstrations in the input context.
Among them, X-ICM~\cite{zhouexploring} explicitly investigated cross-task adaptation by appending textualized demonstrations from seen tasks for relevant unseen tasks~\cite{fan2025test,lee2025ra,sridharricl,wang2025roboflamingo,zhu2024retrieval}.
However, unlike humans who implicitly ``plug'' relevant motor experiences into their execution flow, these methods still rely on fixed structural priors or contextual conditioning, and fail to internalize procedural experience into retrievable and executable model parameters~\cite{chang2025zero,kim2026adaptive,li2026soma}.

\begin{wrapfigure}{r}{0.59\textwidth}
    \vspace{-4mm}
    \centering
    \includegraphics[width=\linewidth]{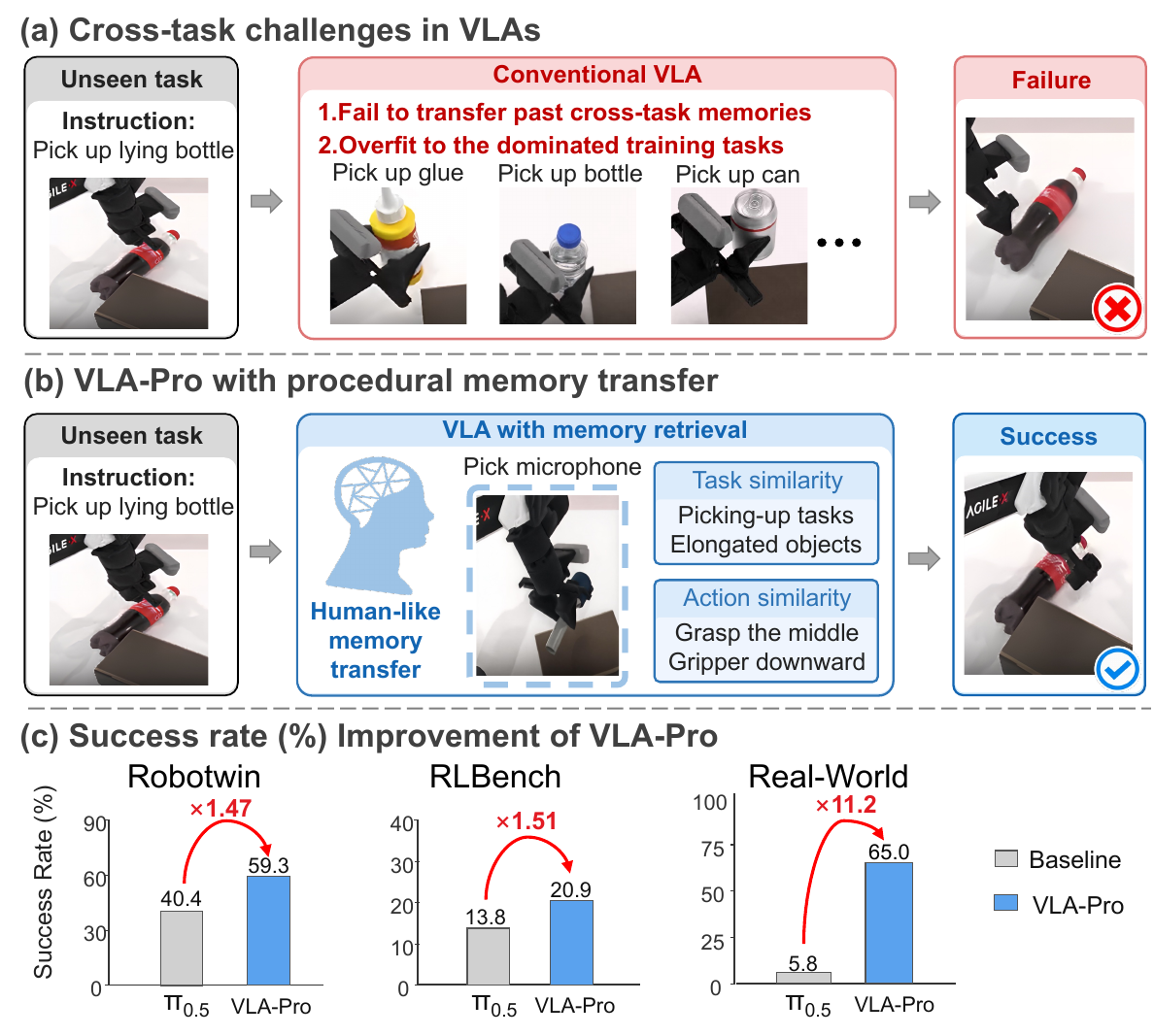}
    \caption{Overview of VLA-Pro.}
    \label{fig:intro}
    \vspace{-4mm}
\end{wrapfigure}


To bridge this gap, we propose VLA-Pro, a plug-and-play framework that transfers procedural memory from the most similar training~(seen) tasks to testing~(unseen) tasks, as shown in Fig.~\ref{fig:intro}.
Specifically, for each seen task, VLA-Pro constructs a procedural memory by pairing a task-specific LoRA adapter with structured procedural states derived from visual observations and language instructions.
These memory entries form a retrievable and reusable memory bank.
At inference time, VLA-Pro extracts the current procedural state and retrieves the top-$k$ most relevant memory entries and fuses their corresponding LoRA adapters based on their semantic similarity.
By dynamically injecting different memories into the pretrained VLA backbone for different execution stages, VLA-Pro can transfer effective procedural experience across tasks.

We conducted comprehensive evaluations of VLA-Pro across a variety of simulation and real-world robotic tasks, including our designed RoboTwin tasks and the RLBench benchmark, as well as real-world robotic scenarios, while employing a cross-task data split to assess the model's generalization on unseen tasks. In the experiments, we compared VLA-Pro with several representative backbone models, including $\pi_{0.5}$, RDT, and X-VLA, as well as existing baseline methods. The results show that VLA-Pro consistently outperforms these baselines across multiple metrics, with particularly strong improvements in cross-task knowledge transfer and decision stability. In specific tasks, VLA-Pro improves the success rate over the corresponding base VLA by approximately 50\%, and in real-world robotic experiments, it achieves an average improvement of approximately 59 percentage points. Moreover, VLA-Pro uses LoRA parameters that occupy minimal storage, and the fusion and switching of parameters are highly efficient, introducing negligible latency without affecting online execution. Ablation analyses further reveal that the performance improves as the retrieved memory becomes more similar to the current task, indicating that scaling the memory bank can further enhance performance. These results collectively demonstrate that VLA-Pro effectively enhances the specialization capability of general-purpose policies, improves cross-task generalization, and maintains stable execution.

In summary, the main contributions of this work are as follows:

\begin{itemize}[leftmargin=*]
    \item We propose VLA-Pro, which enhances VLAs with retrievable and executable procedural memories, enabling efficient cross-task experience transfer to unseen tasks while maintaining policy stability.
    
    \item We analyze the architectural integration of parameterized memories and the efficacy of top-$k$ retrieval, showing that proper memory injection and memory retrieval significantly improve cross-task generalization.
    
    \item We validate the framework on multiple simulation and real-world robotic tasks. VLA-Pro achieves up to 207\% improvement on unseen tasks and demonstrates strong generalization across backbone models, including $\pi_{0.5}$, RDT, and X-VLA.
\end{itemize}

\section{Related Work}

\subsection{Vision-Language-Action Models}
Vision-Language-Action (VLA) models, such as $\pi_0$ ~\cite{black2024pi_0}, $\pi_{0.5}$~\cite{black2025pi_}, OpenVLA ~\cite{kimopenvla}, and X-VLA~\cite{zheng2025x}, unify multimodal perception, linguistic reasoning, and motor control within a single framework, representing a significant advance in embodied intelligence research. These architectures typically encode visual observations and natural language instructions into a shared representation space, leveraging large-scale pre-training to map high-level semantic cues directly to discrete or continuous action outputs\cite{reuss2025flower}. By integrating broad world knowledge from vision-language backbones, VLA agents can execute complex multi-modal tasks and demonstrate strong robustness within familiar training distributions ~\cite{fanlong,nagarajan2018attributes,xia2024kinematic,zhang2025align}. Nonetheless, robust cross-task generalization remains a critical challenge: these models often overfit to trajectory distributions and operational patterns encountered during training, limiting their ability to adapt to novel scenarios without extensive fine-tuning. Our proposed VLA-Pro framework addresses this limitation by mapping structured procedural states to task-specific LoRA memories, enabling cross-task experiences to be parameterized and internalized for reliable zero-shot execution ~\cite{chapman2025queryadapter,fang2025kalm,fu2025mergevla,luo2026coral}.

\subsection{Memory Mechanism for Robotics Control}
Memory mechanisms in robotic control have evolved from simple recurrent hidden states to structured representations that support long-term reasoning and adaptive behavior~\cite{guo2026memeyevisualcentricevaluationframework,xie2026fluxmem,guo2026deepsieve}. Existing approaches~\cite{kuroki2024multi,xie2026dynamicvla} include modularizing parameterized knowledge via skill-based Mixture-of-Experts (MoE) libraries (e.g., AtomicVLA ~\cite{zhang2026atomicvla} ), maintaining dual-stream perceptual-cognitive buffers to store historical context (e.g., MemoryVLA~\cite{shi2025memoryvla}), and employing bio-inspired episodic memory stacks to resolve perceptual ambiguities (e.g., Chameleon~\cite{guo2026chameleon}). Unlike methods that rely on discrete experts or transient contexts, VLA-Pro introduces a retrieval-based procedural memory mechanism. By mapping structured procedural states—incorporating action types and object geometric properties—to task-specific LoRA adapters\cite{qi2024rora}, VLA-Pro internalizes retrieved experiences as modular execution weights, effectively bridging high-level procedural recall with fine-grained motor adaptation~\cite{kumar2025collage,liang2026adaptive}.

\section{Method}
\label{sec:method}

\begin{figure}[thbp]
    \centering
    \includegraphics[width=\linewidth]{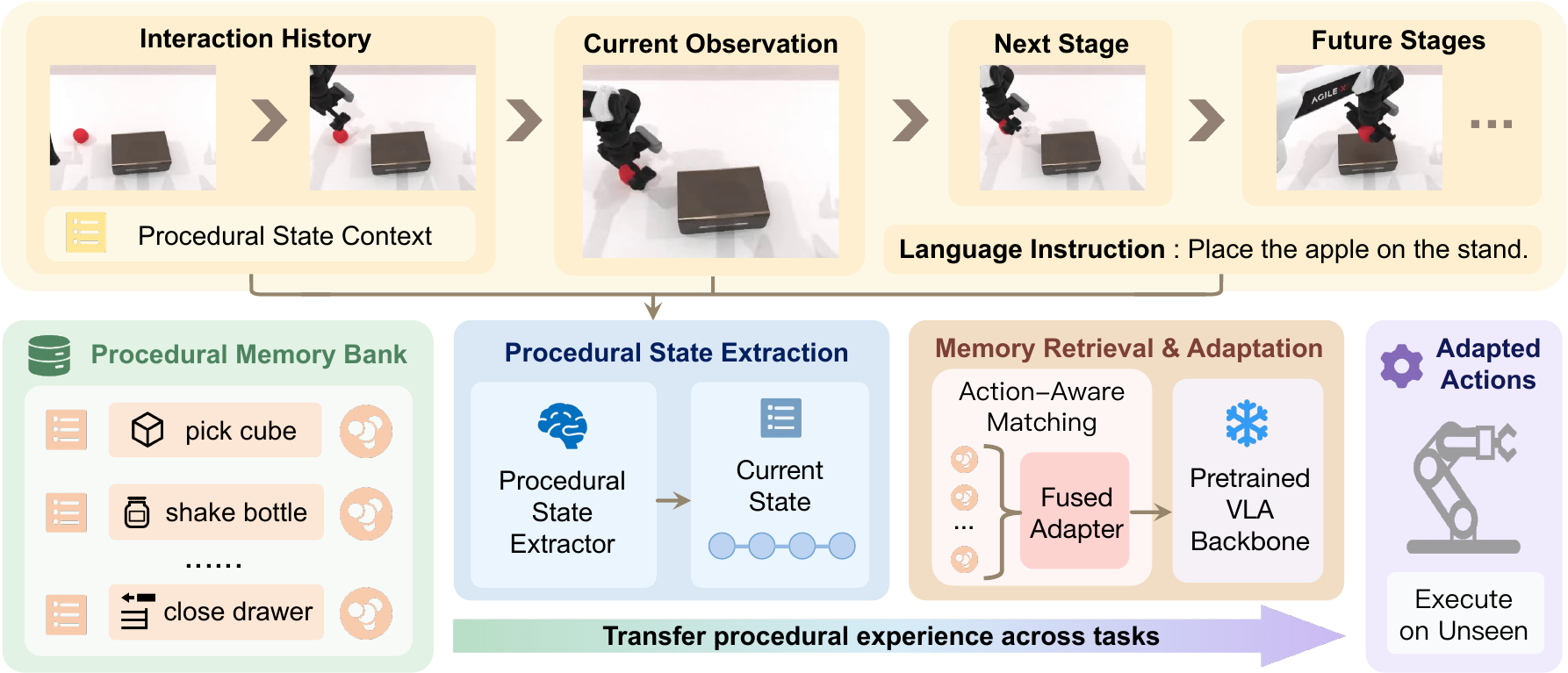}
    \caption{\textbf{VLA-Pro Method Overview.} Task execution consists of sequential stages, each involves the retrieval and integration of procedural memories.
    Given the current multimodal context, VLA-Pro extracts the current procedural state and retrieves relevant memories through action-aware matching.
    Each retrieved memory pairs a procedural state sequence with a task-specific adapter.
    The fused adapter is then applied to the VLA backbone to generate adapted actions for the current stage.
    }
    \label{fig:method_overview}
\end{figure}

To enable cross-task procedural memory transfer in VLAs, we propose VLA-Pro, a plug-and-play framework that adapts relevant memory according to the current procedural state, as illustrated in Figure~\ref{fig:method_overview}. 
Given the current visual observation and language instruction, VLA-Pro first extracts a procedural state that summarizes the ongoing execution stage and uses it to query a procedural memory bank. 
The memory bank is built from previously seen tasks, where each entry stores a task-specific LoRA adapter as a parameterized procedural experience. 
The retrieved LoRA adapters are then fused into the base VLA model, enabling the policy to reuse relevant manipulation patterns while avoiding interference from unrelated task behaviors. 
In this way, VLA-Pro dynamically adapts the model to leverage past memories at runtime, enabling better generalization to unseen tasks.

\subsection{Procedural Memory Storage}
In VLA-Pro, the procedural memory of each seen task $T_i$ is defined as
\begin{equation}
\mathcal{M}_i=(D_i,\Delta\theta_i),
\end{equation}
where $D_i=\{z_i^{(1)},z_i^{(2)},\dots,z_i^{(m_i)}\}$ denotes the sequence of structured procedural states extracted from task $T_i$, and $\Delta\theta_i$ denotes the task-specific LoRA adapter associated with that task. 
This formulation gives procedural memory two complementary roles: $D_i$ specifies the procedural condition under which past experience becomes relevant, while $\Delta\theta_i$ stores the parameterized experience that can be reused for subsequent adaptation.


\begin{wrapfigure}{l}{0.49\linewidth}
\vspace{-0.5em}
\refstepcounter{algorithm}
\label{alg:memory_adaptation}

\hrule
\vspace{0.25em}
\noindent\textbf{Algorithm~\thealgorithm} Pipeline of VLA-Pro
\vspace{0.25em}
\hrule
\vspace{0.35em}

\footnotesize
\begin{algorithmic}[1]
\Statex \textbf{Terminology.}
\Statex Training: \emph{seen tasks}; testing: \emph{unseen tasks}.
\Statex Source memories are constructed from seen tasks.

\Statex \textbf{Base LoRA preparation.}
\State Train base LoRA $\Delta\theta_{\mathrm{base}}$ on seen tasks.
\State Use the base-LoRA policy as the baseline.

\Statex \textbf{Offline memory construction.}
\For{each selected seen task $T_i$}
    \State Extract procedural states $D_i$.
    \State Fine-tune from $\Delta\theta_{\mathrm{base}}$ to obtain task LoRA $\Delta\theta_i$.
    \State Store memory $\mathcal{M}_i=(D_i,\Delta\theta_i)$.
\EndFor

\Statex \textbf{Online adaptation on unseen tasks.}
\For{each action chunk $t$}
    \State Extract current state $\hat{z}_t$.
    \State Retrieve source memories indexed by $\mathcal{K}_t$.
    \State Compute fusion weights $\{\alpha_j\}_{j\in\mathcal{K}_t}$.
    \State Merge retrieved task LoRAs into $\Delta\theta_t^{\mathrm{fused}}$.
    \State Load $\Delta\theta_t^{\mathrm{fused}}$ on top of base parameters.
    \State Execute the chunk and unload $\Delta\theta_t^{\mathrm{fused}}$.
\EndFor
\end{algorithmic}

\vspace{0.25em}
\hrule
\vspace{-0.5em}
\end{wrapfigure}

Instead of directly using raw observations or free-form language descriptions, we abstract each execution stage into a structured procedural state
\begin{equation}
z=\{a,o,e,p\},
\end{equation}
where $a$, $o$, $e$, and $p$ correspond to the fields of the current action type, object-related geometric property, end-effector orientation, and target interaction point, respectively. 
This compact representation makes procedural experience more indexable and retrievable.

We first perform joint LoRA fine-tuning on seen tasks to obtain a shared base adapter $\Delta\theta_{\mathrm{base}}$, which captures general manipulation knowledge across all training tasks. 
Starting from $\Delta\theta_{\mathrm{base}}$, we then fine-tune a separate adapter for each task $T_i$ with its own demonstrations, resulting in the task-specific adapter $\Delta\theta_i$. 
During inference, the retrieved task-specific adapters are fused and applied to the same pretrained VLA backbone, preserving modularity while avoiding interference.


\subsection{Cross-Task Memory Retrieval}
Given the current observation and the full language instruction, the goal of cross-task memory retrieval is to identify the most relevant procedural memories from previously seen tasks for the current execution stage. 
Instead of directly retrieving from raw images or free-form language descriptions, VLA-Pro performs retrieval over structured procedural states, which enables more precise and interpretable memory addressing under fine-grained differences in actions and interaction geometry. 
As summarized in Algorithm~\ref{alg:memory_adaptation}, the retrieved top-$k$ memories are then used for memory merging in the next stage.

\subsubsection{Current Procedural State Extraction}
Before the model generates an action chunk, we extract the current procedural state as the retrieval query:
\begin{equation}
\hat{z}_t = \mathcal{E}(O_t, l, H_{<t}),
\end{equation}
where $O_t$ denotes the current observation, $l$ denotes the full task instruction, $H_{<t}$ denotes the interaction history before step $t$, and $\mathcal{E}(\cdot)$ denotes the procedural state extractor.

VLA-Pro uses a strong vision-language model as the procedural state extractor to infer the current execution stage from these inputs. The interaction history $H_{<t}$ consists of previous images and their corresponding structured procedural states, with redundant repeated items removed, which helps the model capture the temporal order of subtasks and better predict the following stage. The extraction is guided by a structured schema consistent with source memories, making the retrieval query more explicit and interpretable than raw visual or textual embeddings. The details of the prompt for procedural state extraction and the procedural state schema are provided in Appendix~\ref{app:storage_retrieval}.

\subsubsection{Task-Wise Memory Retrieval with Action-Aware Procedural Matching}
Each source memory is indexed by its procedural state sequence $D_i$. Given the current query state $\hat{z}_t$, we compare it with all states in $D_i$ and define the task-level relevance by the best-matched state. For efficiency, the textual procedural states are pre-encoded and cached before online retrieval:
\begin{equation}
s_i = \max_{m \in \{1,\dots,m_i\}} \mathrm{Sim}\!\left(\hat{z}_t, z_i^{(m)}\right).
\end{equation}

To capture action-dependent procedural requirements, we adopt an action-aware matching mechanism. For a procedural state $z=\{a,o,e,p\}$, the action field $a$ determines the relative weights of $o$, $e$, and $p$, so different action categories emphasize different procedural attributes. Formally, we write
\begin{equation}
\operatorname{Sim}\!\left(\hat{z}_t, z_i^{(m)}\right)
=
\sum_{f \in \{a,o,e,p\}} w_f(\hat{a}_t) \, \operatorname{Sim}_f\!\left(\hat{z}_{t,f}, z_{i,f}^{(m)}\right),
\end{equation}
where $\hat{a}_t$ denotes the action field of $\hat{z}_t$, $\operatorname{Sim}_f(\cdot,\cdot)$ denotes the cosine similarity between the text embeddings of field $f$, and $w_f(\hat{a}_t)$ is the action-conditioned weight. The detailed procedural-state schema and action-aware weight settings are provided in Appendix~\ref{app:state_schema_matching}. This design is intended to adapt the matching process to object attributes that are strongly coupled with the current action.

After computing the task-level similarities for all source memories, we retain the top-$k$ most relevant source memories and denote their index set by $\mathcal{K}_t$. These retrieved procedural memories and their similarities are then passed to the next stage for memory merge and adaptation.

\subsection{Memory Merging and Adaptation}

Starting from the retrieved top-$k$ source memories and their task-level similarities, VLA-Pro converts these similarities into fusion coefficients for memory composition. Specifically, we apply a Softmax function over the similarity scores of the retrieved tasks to obtain normalized fusion weights $\{\alpha_j\}_{j\in\mathcal{K}_t}$. These coefficients indicate the relative contribution of different retrieved memories under the current procedural state. Based on these coefficients, the corresponding task-specific LoRA adapters are fused through parameter-wise weighted summation:
\begin{equation}
\Delta\theta_t^{\mathrm{fused}}=\sum_{j\in\mathcal{K}_t}\alpha_j \Delta\theta_j,
\end{equation}
where $\mathcal{K}_t$ denotes the index set of the retrieved top-$k$ source memories, and $\Delta\theta_j$ denotes the task-specific LoRA adapter associated with the retrieved task indexed by $j$. In this way, the retrieved procedural memories are merged into a single adapter tailored to the current execution stage.

Rather than modifying the pretrained model directly, VLA-Pro loads the fused adapter onto the pretrained VLA model parameters to form a temporary policy for the current action chunk. After the current chunk is executed, the fused adapter is unloaded and the model is restored to the original base state. The system then proceeds to the next execution stage and repeats this process based on the updated interaction history.

\section{Experimental Setup}
\label{sec:setup}
We organize our experiments to address the following key research questions:
\begin{itemize}[leftmargin=*]
    \item \textbf{RQ1:} Can VLA-Pro effectively improve the performance on unseen tasks?
    \item \textbf{RQ2:} What roles do different parameter components play in cross-task generalization?
    \item \textbf{RQ3:} How does the retrieval accuracy of procedural memory affect the overall model performance?
    \item \textbf{RQ4:} How does the number of retrieved memories (i.e., top-$k$ selection) influence the model’s generalization and decision-making stability?
\end{itemize}
This section presents the dataset construction, model selection, and real-world experimental setup. More details on task design and implementation are provided in Appendices~\ref{app:task_suite} and~\ref{app:implementation_details}.

\subsection{Dataset Construction and Evaluation}
We evaluate VLA-Pro across two simulated benchmarks, RoboTwin~\cite{chen2025robotwin} and RLBench~\cite{james2020rlbench}, as well as in real-world manipulation scenarios.
In all settings, we adopt a strict separation between training (seen) and testing (unseen) tasks to rigorously test cross-task generalization.
Task success rates are computed using the officially provided success checker for the simulated environment and through manual verification in real-world experiments. 

\noindent\textbf{RoboTwin}
Our RoboTwin training set consists of 2 tasks from the official RoboTwin benchmark and 6 newly designed tasks, while the evaluation set contains 9 newly designed held-out tasks based on the RoboTwin environment~(see Fig.~\ref{fig:robotwin_data}).
The testing tasks share relevant objects or actions with the training tasks, but are not identical to any training task. 
This task suite is designed to assess the cross-task transfer ability of the proposed method.
Following the data collection protocol of the original RoboTwin benchmark, 50 trajectories are collected for training on each seen task.

\noindent\textbf{RLBench}
We follow the cross-task generalization evaluation protocol established by X-ICM\citep{zhouexploring}. 
The X-ICM dataset provides 100 expert demonstrations for each of the 18 designated training~(seen) tasks. We train the base LoRA adapter on all 18 seen tasks, and select 8 of them as the source tasks for constructing procedural memories. Performance is evaluated on all zero-shot tasks.

\begin{figure*}[htbp] 
    \centering
    \includegraphics[width=0.8\textwidth]{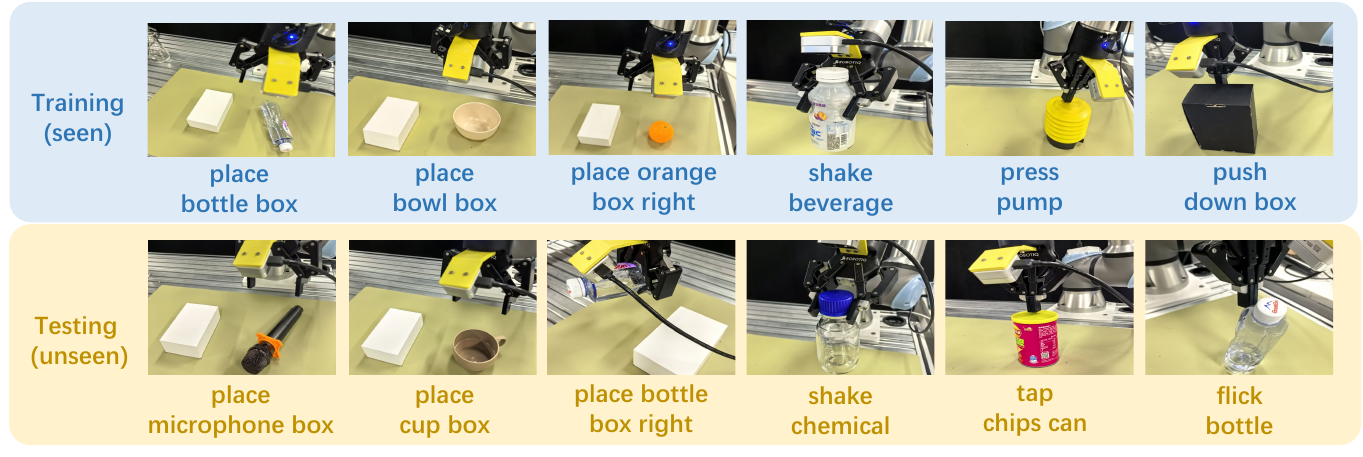}
    \caption{Overview of our real-world experimental setup. 6 training tasks and 6 corresponding test tasks designed for evaluating the model's performance in real-world manipulation.}
    \label{fig:real_world_setup}
\end{figure*}

\paragraph{Real-world Experiments}
The real-world experiments are conducted on a UR7e robot arm equipped with a Robotiq 2F-85 gripper and a wrist-mounted RealSense D405 camera, which provides single-view visual observations. 
The policy operates in the end-effector pose action space. 
The real-world experiment consists of 6 training tasks and 6 corresponding held-out testing tasks, as shown in Fig.~\ref{fig:real_world_setup}. 
$\pi_{0.5}$ is used as the backbone to verify the effectiveness of the proposed method. 
For each training task, 50 demonstrations are manually collected, and each testing task is evaluated over 20 trials.

\subsection{Models}
\paragraph{Backbones and Baselines} For RoboTwin, the VLA-Pro framework is instantiated on three different backbones: X-VLA\citep{zheng2025x}, RDT\citep{liu2024rdt}, and $\pi_{0.5}$\citep{black2025pi_}. 
For each backbone, its pretrained checkpoint independent of RoboTwin is used as the base model for LoRA fine-tuning.
The baseline is established by training a base LoRA adapter across all 8 RoboTwin training tasks.
Building upon this, VLA-Pro further trains separate task-specific LoRA adapters for each training task to encapsulate procedural memories.
For RLBench, $\pi_{0.5}$ is used as the backbone, and the baseline is trained jointly on all 18 training tasks. 
Our method is compared with RDT, $\pi_{0.5}$, and AtomicVLA\citep{zhang2026atomicvla}.

\paragraph{Procedural Memories Retrieval} 
During inference, Gemini-3-Flash is used as the procedural state extractor, and OpenAI's text-embedding-3-large is used for Action-Aware Procedural Matching. 
For RoboTwin, we set the number of retrieved memory candidates to $k=1$, as the limited number of training tasks facilitates more precise single-candidate matching. 
For RLBench, where the training tasks are more diverse, we employ a top-$k$ selection of $k=2$. 
Additional training hyperparameters and implementation details are provided in Appendix~\ref{app:implementation_details}.

\section{Experimental Results}
\label{sec:results}
\subsection{Overall Results}


\begin{table}[htbp]
\centering
\small
\setlength{\tabcolsep}{0pt}
\renewcommand{\arraystretch}{1.12}
\caption{Success rate (\%) on the RoboTwin benchmark. 
Each entry presents the success rate for one task over 30 trials, and Avg. denotes the mean across all tasks.}
\label{tab:robotwin}
\begin{tabular}{
>{\raggedright\arraybackslash}m{1.6cm}
>{\centering\arraybackslash}m{1.5cm}@{\hspace{4pt}}|
>{\centering\arraybackslash}m{1.2cm}
>{\centering\arraybackslash}m{1.2cm}
>{\centering\arraybackslash}m{1.2cm}
>{\centering\arraybackslash}m{1.2cm}
>{\centering\arraybackslash}m{1.2cm}
>{\centering\arraybackslash}m{1.2cm}
>{\centering\arraybackslash}m{1.2cm}
>{\centering\arraybackslash}m{1.2cm}
>{\centering\arraybackslash}m{1.2cm}
>{\centering\arraybackslash}m{1.7cm}}
\toprule
\multicolumn{2}{c|}{\textbf{Method}}
& \makecell{apple\\stand}
& \makecell{rcube\\stand}
& \makecell{cup\\stand}
& \makecell{lying\\bottle}
& \makecell{upright\\bottle}
& \makecell{bell\\behind}
& \makecell{close\\micro.}
& \makecell{close\\cab.}
& \makecell{stack\\bowls}
& Avg. \\
\midrule
\multirow{2}{*}{\centering {X-VLA}\citep{zheng2025x}}
& base
& 0.0 & 3.3 & 23.3 & 0.0 & 30.0 & 0.0 & 43.3 & 53.3 & 0.0 & 17.0 \\
& VLA-Pro
& \cellcolor{blue!12}3.3 & \cellcolor{blue!12}23.3 & \cellcolor{blue!12}46.7 & \cellcolor{blue!12}0.0 & \cellcolor{blue!12}46.7 & \cellcolor{blue!12}0.0 & \cellcolor{blue!12}50.0 & \cellcolor{blue!12}96.7 & \cellcolor{blue!12}3.3 & \cellcolor{blue!12}\textbf{30.0}{\color{green!50!black}\scriptsize(76\%$\uparrow$)} \\
\midrule
\multirow{2}{*}{\centering RDT\citep{liu2024rdt}}
& base
& 16.7 & 13.3 & 30.0 & 13.3 & 0.0 & 0.0 & 10.0 & 13.3 & 3.3 & 11.1 \\
& VLA-Pro
& \cellcolor{blue!12}26.7 & \cellcolor{blue!12}23.3 & \cellcolor{blue!12}40.0 & \cellcolor{blue!12}16.7 & \cellcolor{blue!12}6.7 & \cellcolor{blue!12}6.7 & \cellcolor{blue!12}70.0 & \cellcolor{blue!12}80.0 & \cellcolor{blue!12}36.7 & \cellcolor{blue!12}\textbf{34.1}{\color{green!50!black}\scriptsize(207\%$\uparrow$)} \\
\midrule
\multirow{2}{*}{\centering {$\pi_{0.5}$\citep{black2025pi_}}}
& base
& 20.0 & 6.7 & 16.7 & 73.3 & 46.7 & 0.0 & 96.7 & 76.7 & 26.7 & 40.4 \\
& VLA-Pro
& \cellcolor{blue!12}33.3 & \cellcolor{blue!12}16.7 & \cellcolor{blue!12}33.3 & \cellcolor{blue!12}86.7 & \cellcolor{blue!12}66.7 & \cellcolor{blue!12}100.0 & \cellcolor{blue!12}70.0 & \cellcolor{blue!12}96.7 & \cellcolor{blue!12}30.0 & \cellcolor{blue!12}\textbf{59.3}{\color{green!50!black}\scriptsize(47\%$\uparrow$)} \\
\bottomrule
\end{tabular}
\end{table}

\noindent\textbf{Results on RoboTwin.}
As shown in Table~\ref{tab:robotwin}, VLA-Pro yields consistent performance improvement across all three backbones, elevating the average success rate of X-VLA from 17.0\% to 30.0\%, RDT from 11.1\% to 34.1\%, and $\pi_{0.5}$ from 40.4\% to 59.3\%. 
This demonstrates that the proposed procedural memory mechanism can generalize across different VLA backbones. 
Instead of relying on a single mixed-task LoRA, VLA-Pro uses procedural state extraction and Action-Aware Procedural Matching to transfer relevant manipulation patterns. 
For example, on the \textit{place\_bell\_behind} task, the $\pi_{0.5}$ baseline obtains 0.0\%, while VLA-Pro achieves 100.0\%, showing that dynamic memory retrieval helps the model avoid overusing the dominant ``place-on-stand'' behavior and switch to the correct spatial-relation memory.


\begin{table}[htbp]
\centering
\small
\setlength{\tabcolsep}{0pt}
\renewcommand{\arraystretch}{1.12}
\caption{Success rate (\%) on the RLBench benchmark. 
Each entry reports the success rate for one zero-shot task over 25 trials. 
The table presents the 9 testing tasks with non-zero success for at least one method. 
Avg. denotes the mean success rate.
VLA-Pro is built upon $\pi_{0.5}$ in this experiment.
}
\label{tab:rlbench}
\begin{tabular}{
>{\raggedright\arraybackslash}m{2.4cm}|
>{\centering\arraybackslash}m{1.3cm}
>{\centering\arraybackslash}m{1.3cm}
>{\centering\arraybackslash}m{1.3cm}
>{\centering\arraybackslash}m{1.3cm}
>{\centering\arraybackslash}m{1.3cm}
>{\centering\arraybackslash}m{1.5cm}
>{\centering\arraybackslash}m{1.3cm}
>{\centering\arraybackslash}m{1.3cm}
>{\centering\arraybackslash}m{1.3cm}
>{\centering\arraybackslash}m{1.5cm}}
\toprule
\makecell[c]{\textbf{Method}}
& \makecell{close\\fridge}
& \makecell{laptop\\lid}
& \makecell{turn\\oven on}
& \makecell{toilet\\seat}
& \makecell{water\\plants}
& \makecell{close\\microwave}
& \makecell{take\\usb out}
& \makecell{take\\lid off}
& \makecell{beat\\the buzz}
& Avg. \\
\midrule
RDT\citep{liu2024rdt} & 20.0 & 12.0 & 0.0 & 16.0 & 0.0 & 12.0 & 32.0 & 0.0 & 0.0 & 10.2 \\
$\pi_{0.5}$\citep{black2025pi_} & 24.0 & 0.0 & 0.0 & 28.0 & 0.0 & 0.0 & 72.0 & 0.0 & 0.0 & 13.8 \\
AtomicVLA\citep{zhang2026atomicvla} & 64.0 & 4.0 & 12.0 & 16.0 & 0.0 & 0.0 & 36.0 & 0.0 & 0.0 & 14.7 \\
\rowcolor{blue!12}
VLA-Pro\_{$\pi_{0.5}$} & 40.0 & 4.0 & 8.0 & 32.0 & 4.0 & 4.0 & 88.0 & 4.0 & 4.0 & \textbf{20.9}{\color{green!50!black}\scriptsize(51\%$\uparrow$)} \\
\bottomrule
\end{tabular}
\end{table}
\FloatBarrier

\noindent\textbf{Results on RLBench.}
As shown in Table~\ref{tab:rlbench}, VLA-Pro achieves the best average success rate over 9 zero-shot tasks in RLBench.
It outperforms RDT, AtomicVLA, and the $\pi_{0.5}$ baseline by 10.7, 6.2, and 7.1 percentage points, respectively.
This result shows that, with a richer base LoRA adapter trained on all 18 seen tasks, VLA-Pro can still selectively reuse useful procedural experiences to further improve zero-shot performance. 

\begin{figure}[H]
    \centering
    \captionsetup[subfigure]{skip=1pt}

    \begin{subfigure}[t]{0.48\linewidth}
        \centering
        \includegraphics[width=\linewidth,height=3.5cm,keepaspectratio]{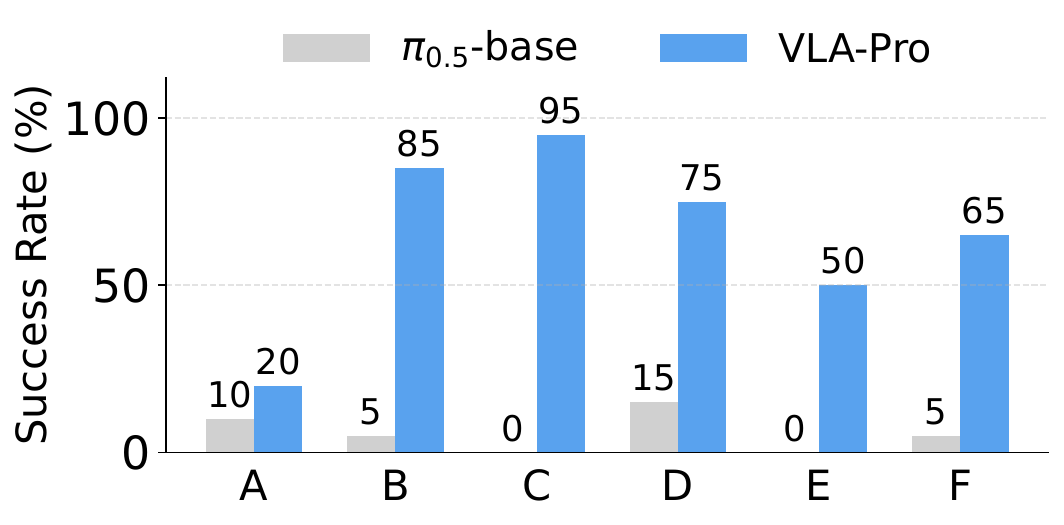}
        \caption{Success rates on six real-world manipulation tasks. A -- F denote the tasks in the same order as Figure~\ref{fig:real_world_setup}.}
        \label{fig:realworld_results}
    \end{subfigure}
    \hfill
    \begin{subfigure}[t]{0.48\linewidth}
        \centering
        \includegraphics[width=\linewidth,height=4cm,keepaspectratio]{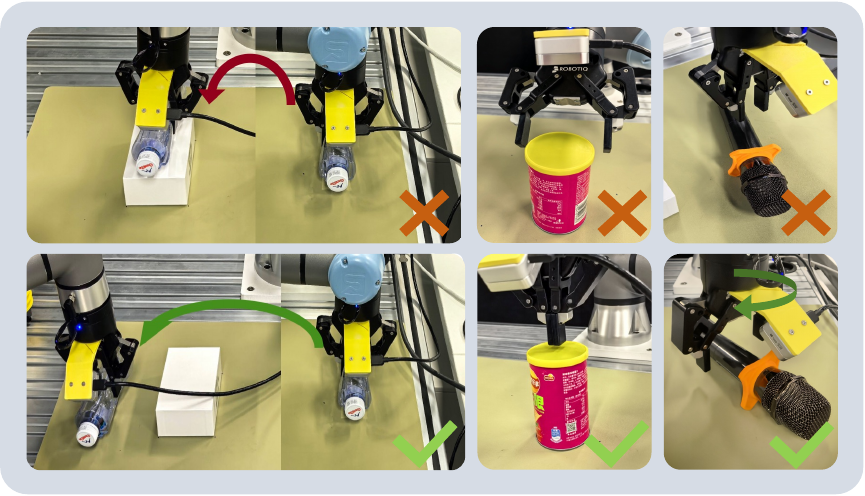}
        \caption{Real-world task comparison between the baseline~(upper row) and VLA-Pro~(lower row).}
        \label{fig:realworld_demo}
    \end{subfigure}

    \caption{\textbf{Real-world manipulation results.} Quantitative success rates and qualitative execution examples on held-out real-world tasks, comparing the baseline with VLA-Pro.}
    \label{fig:realworld_results_combined}
\end{figure}
\FloatBarrier

\noindent\textbf{Real-world Experimental Results.}
Figure~\ref{fig:realworld_results_combined} shows that VLA-Pro substantially improves real-world cross-task generalization. As shown in Figure~\ref{fig:realworld_results}, using $\pi_{0.5}$ as the backbone, VLA-Pro increases the average success rate from 5.8\% to 65.0\% over six held-out tasks. 
Figure~\ref{fig:realworld_demo} further illustrates several examples showing that procedural memory improves VLA's performance. 
For the \textit{place bottle box right} task, the baseline tends to follow dominant placement patterns from training tasks, while VLA-Pro retrieves relevant memory to guide the correct actions. 
This shows its ability to mitigate high-frequency execution biases. 
For \textit{tap the chips can} and \textit{place mic on box}, VLA-Pro retrieves related object-level patterns, demonstrating reusable procedural transfer in real-world tasks.

\subsection{Procedural Memory Parameter Integration}
\label{sec:rq2}

To answer RQ2, we select five pick-and-place tasks from RoboTwin. 
The selection ensures that each unseen task has a unique corresponding most similar seen task~(see Appendix~\ref{app:robotwin_tasks}), which enables a fair comparison for investigating parameter transfer effects.

\begin{figure}[ht]
    \centering
    \begin{minipage}[t]{0.46\linewidth}
        \centering
        \includegraphics[width=\linewidth]{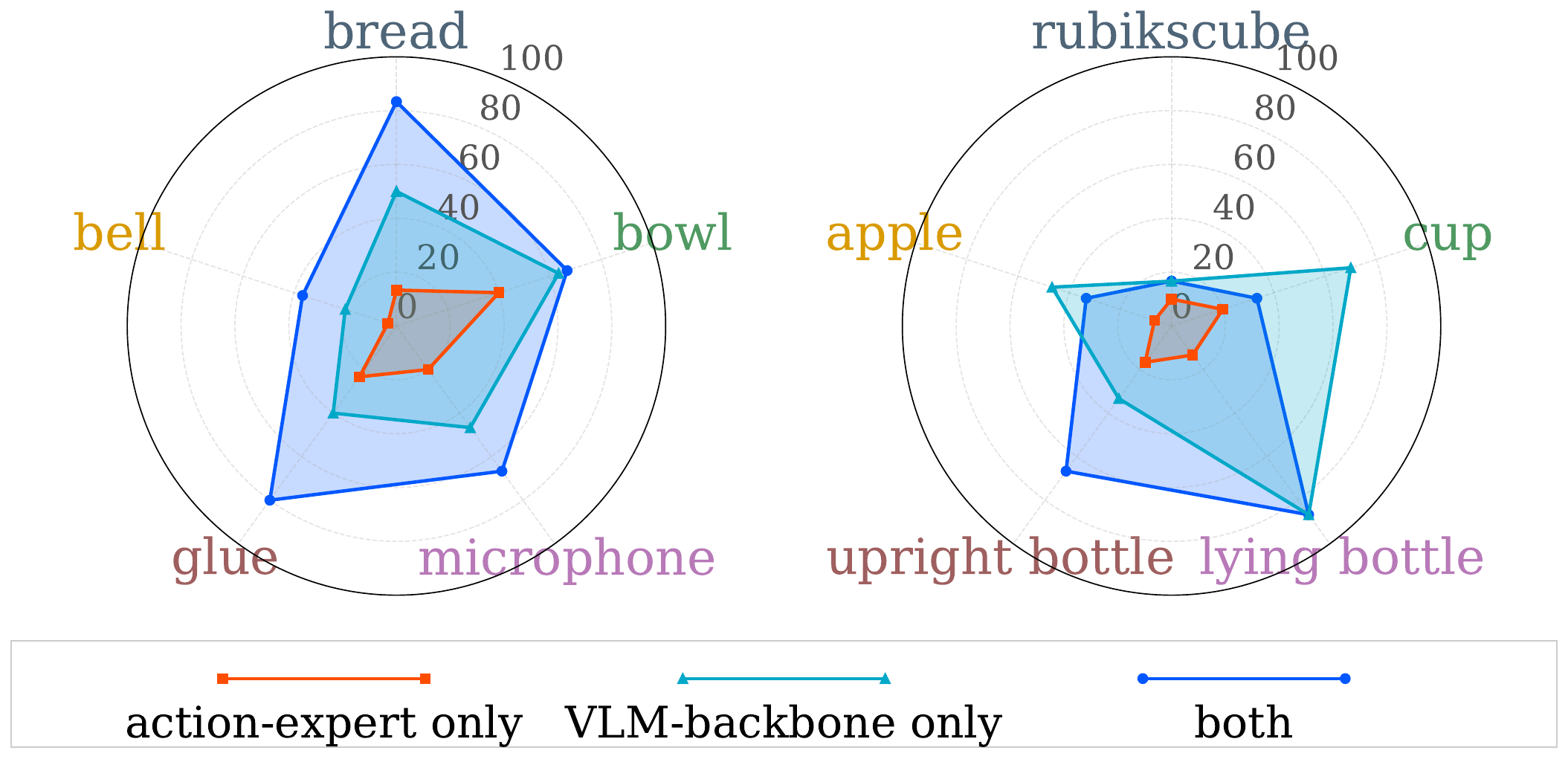}
        \caption{Effect of different parameter components in VLA-Pro with $\pi_{0.5}$ backbone on RoboTwin. The left and right radar charts show five seen and five unseen tasks, respectively.}
        \label{fig:rq2_radar}
    \end{minipage}
    \hfill
    \begin{minipage}[t]{0.50\linewidth}
        \centering
        \includegraphics[width=\linewidth]{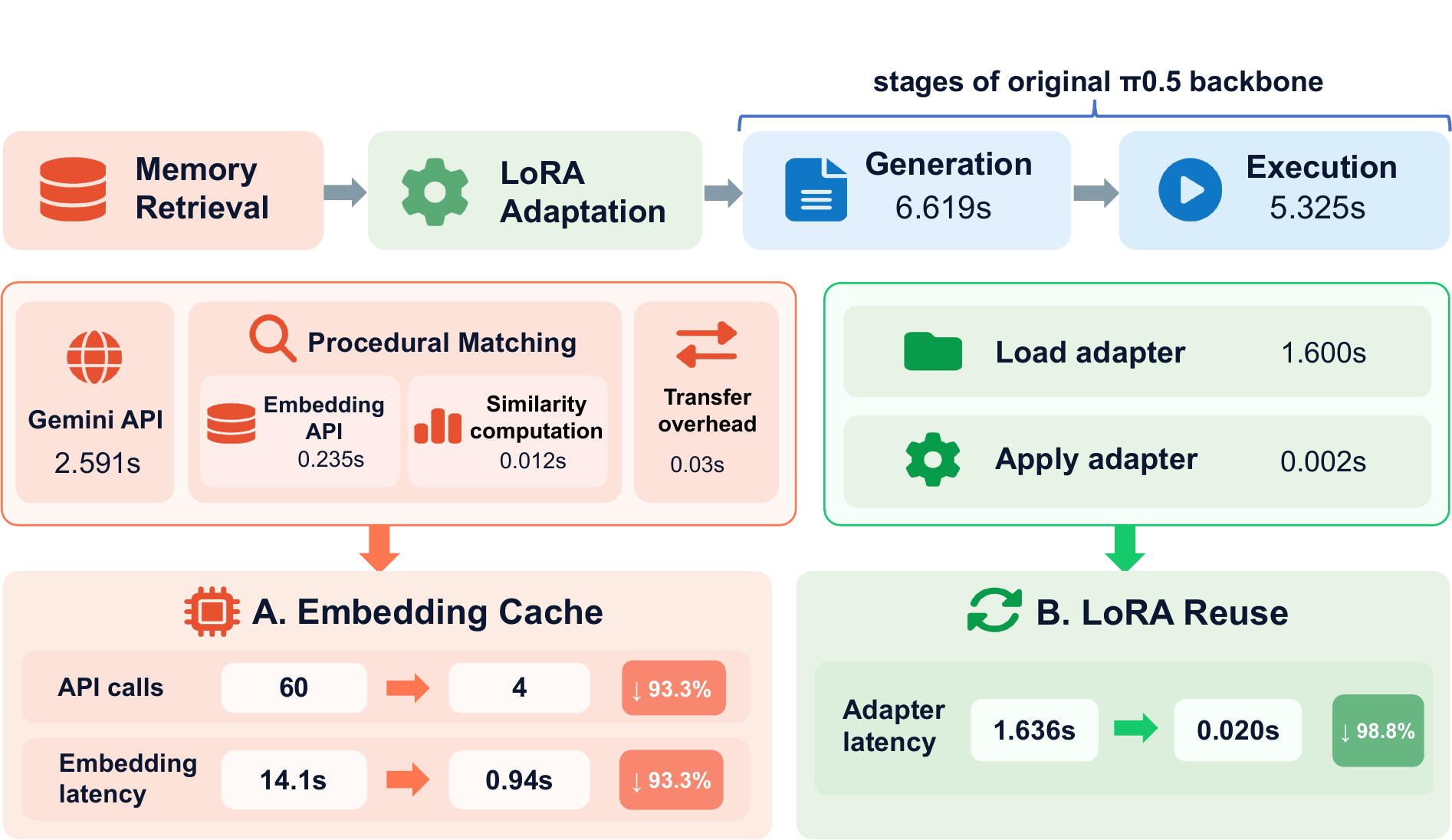}
        \caption{Runtime analysis of VLA-Pro, showing the time latency introduced by the additional modules on top of the $\pi_{0.5}$ backbone.}
        \label{fig:rq2_latency}
    \end{minipage}
\end{figure}

\noindent
Typical VLA models consist of a VLM-backbone for vision-language understanding and an action-expert for action generation. 
To explore their roles in memory transfer, we separate the parameterized procedural experience into VLM-backbone and action-expert components and evaluate them on seen and unseen tasks. 
As shown in Fig.~\ref{fig:rq2_radar}, these components exhibit complementary behaviors: using both achieves the strongest performance on seen tasks, indicating that jointly adapting vision-language reasoning and action execution improves task-specific behavior.

\vspace{0.5em}
\noindent
In contrast, on unseen tasks, generalization gains with VLA-Pro are task-dependent.
For tasks that are highly similar to the dominant patterns in the training set~(e.g., pick up upright bottle), adapting the entire parameter set ensures the highest execution precision by leveraging specialized action-expert knowledge. 
However, for tasks requiring significant zero-shot generalization~(e.g., pick up cup and apple), the VLM-backbone adaptation proves more effective.
This suggests that the VLM-backbone's strength lies in its ability to extract and apply high-level semantic and physical reasoning cues to novel scenarios.

Fig.~\ref{fig:rq2_latency} shows the runtime overhead introduced by VLA-Pro. 
VLA-Pro improves efficiency by caching text embeddings of source memories and reusing loaded LoRA adapters during online adaptation. 
In our measurement, the average latency of a single Gemini API call is 2.591s and that of each text-embedding API call is 0.235s, while cached retrieval and LoRA reuse substantially reduce repeated computation and loading costs.
As a result, the extra latency introduced by VLA-Pro is relatively small compared with the total time spent on action-chunk generation and execution.

\subsection{Retrieval Effectiveness Analysis}
To answer RQ3, we evaluate the impact of different retrievers and retrieved task similarity on the performance improvement of VLA-Pro, respectively. 

\begin{figure}[htbp]
    \centering
    \captionsetup[subfigure]{labelformat=parens,labelsep=none}

    \begin{minipage}[t]{0.58\linewidth}
        \vspace{0pt}
        \centering

        \begin{minipage}[t][4.65cm][t]{\linewidth}
            \vspace{0pt}
            \centering

            \begin{subfigure}[t]{0.39\linewidth}
                \centering
                \makebox[\linewidth][c]{%
                    \includegraphics[height=4cm]{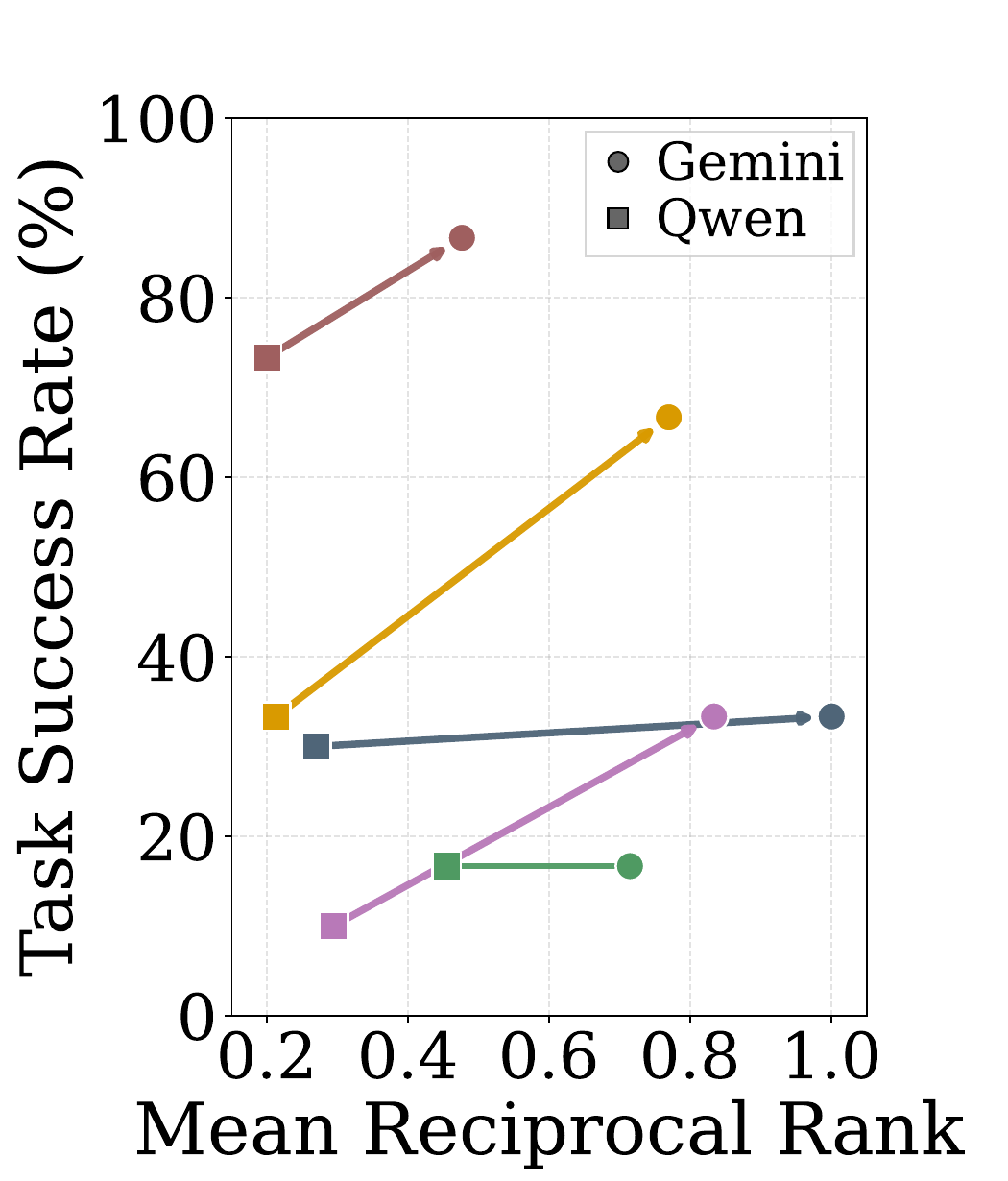}
                }
                \caption{}
                \label{fig:retrieval_performance_left}
            \end{subfigure}
            \hfill
            \begin{subfigure}[t]{0.57\linewidth}
                \centering
                \makebox[\linewidth][c]{%
                    \includegraphics[height=3.8cm]{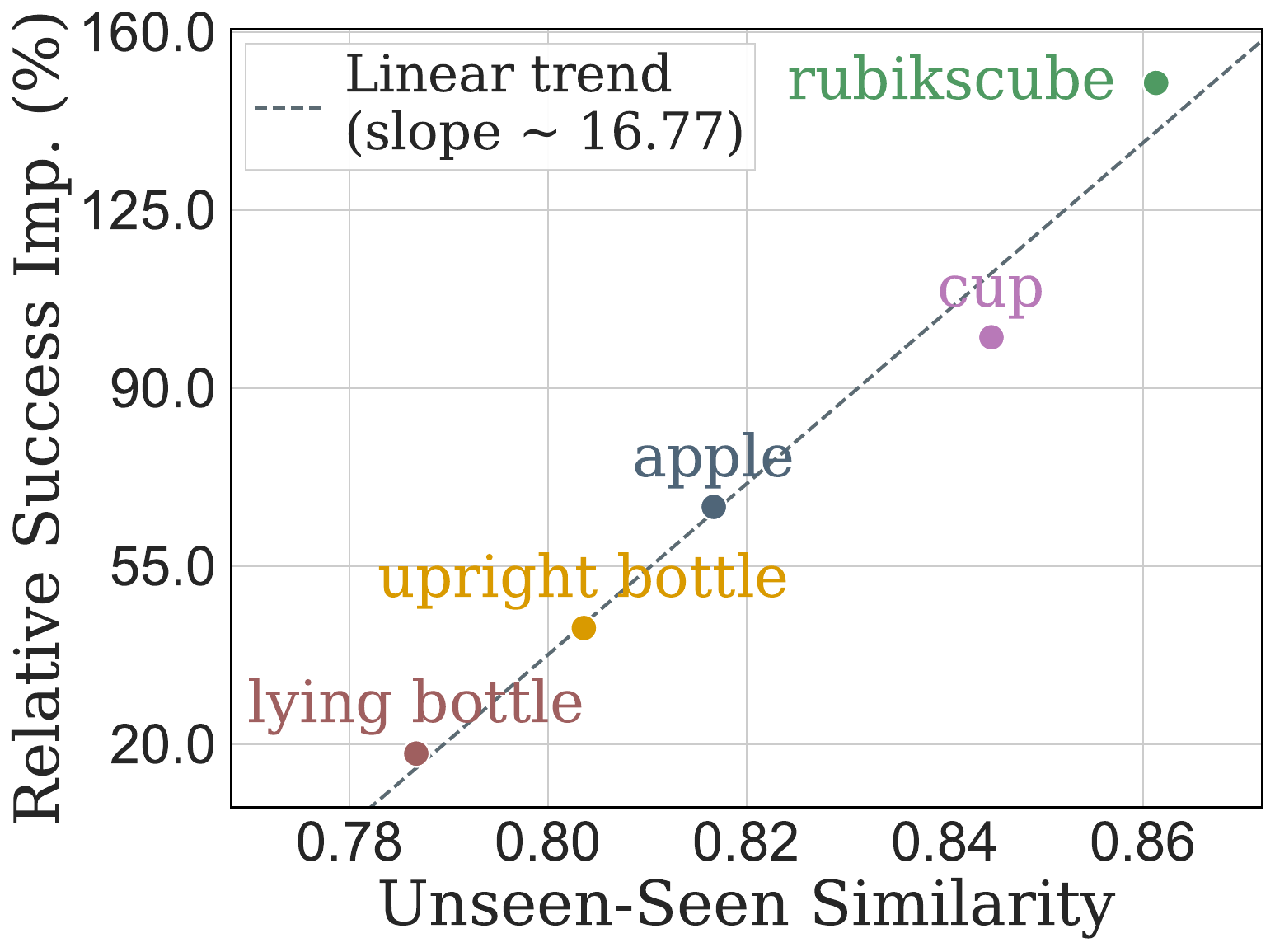}
                }
                \caption{}
                \label{fig:retrieval_performance_right}
            \end{subfigure}

        \end{minipage}

        \caption{Analysis of retrieval performance vs. task success rate. 
        (a) Procedural state extraction accuracy measured by MRR. 
        (b) Correlation between Unseen--Seen task similarity and transfer gain. 
        }
        \label{fig:retrieval_performance}
    \end{minipage}
    \hfill
    \begin{minipage}[t]{0.38\linewidth}
        \vspace{0pt}
        \centering

        \begin{minipage}[t][4.65cm][t]{\linewidth}
            \vspace{0pt}
            \centering
            \includegraphics[height=4.8cm]{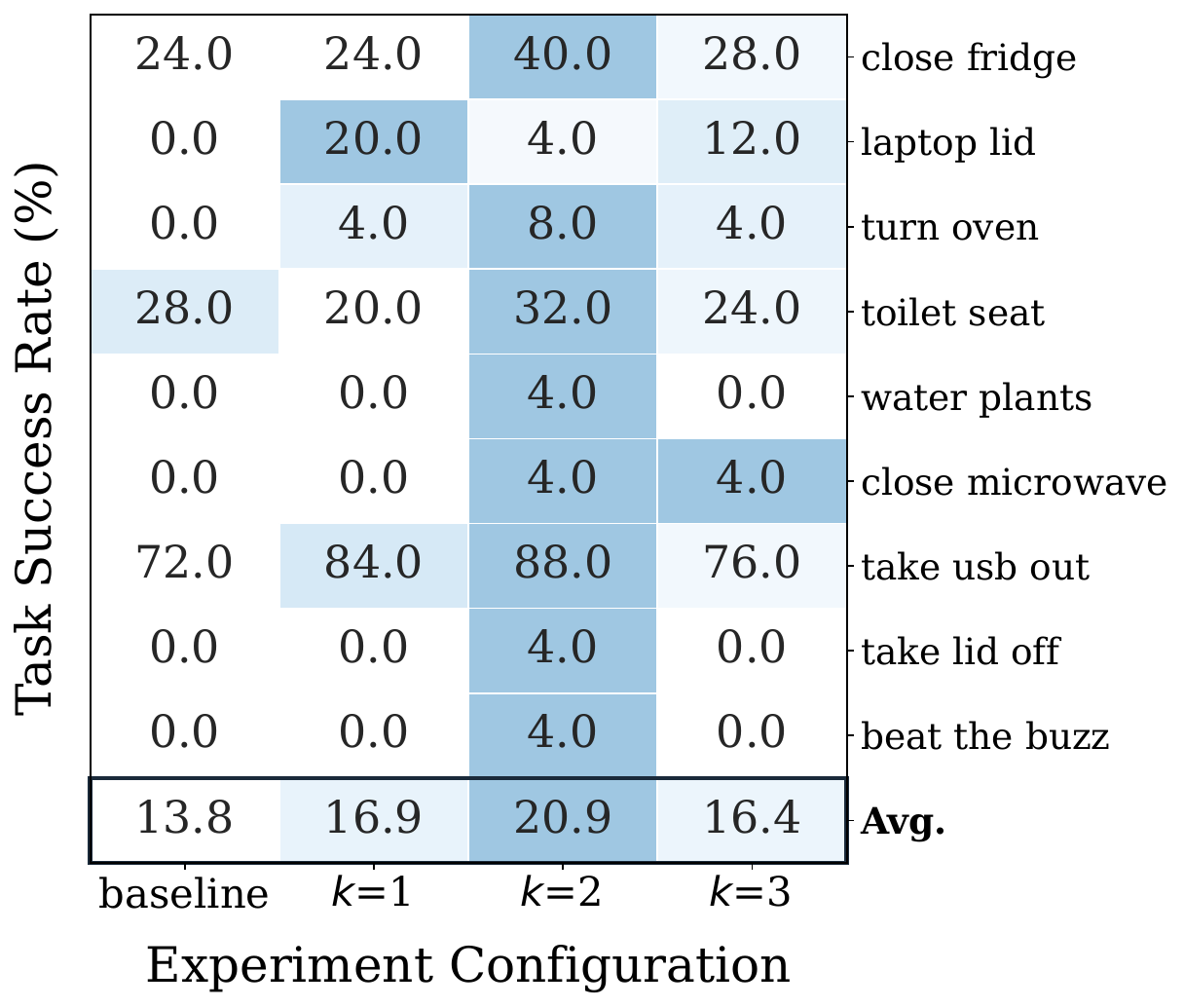}
        \end{minipage}

        \caption{Effect of different top-$k$ retrieval settings on RLBench testing tasks. Darker colors indicate higher task success rates.}
        \label{fig:topk_analysis}
    \end{minipage}

\end{figure}

\noindent\textbf{Impact of retrievers.}
We first evaluate the effect of the selected retrievers, including Gemini-3-Flash and Qwen2.5-VL-7B, with the embedding model and the VLA policy fixed. 
We compare these retrievers based on mean reciprocal rank~(MRR) computed by matching the VLM-extracted state against the predefined procedural states in the memory bank.
As shown in Figure~\ref{fig:retrieval_performance_left}, Gemini achieves higher MRR than Qwen across tasks.
This indicates Gemini achieves more accurate retrieval, and further leads to higher physical execution success rates.

\noindent\textbf{Impact of task similarity.}
We quantify the correlation between VLA-Pro's performance and task similarity by measuring the retriever's similarity score between each unseen task and its corresponding source memories.
As shown in Figure~\ref{fig:retrieval_performance_right}, tasks with higher similarity generally obtain larger performance gains. 
This result indicates that more similar source memories provide more transferable procedural experience, leading to better execution performance on unseen tasks.

\subsection{Effect of Top-$k$ Memory Merging}

This subsection investigates RQ4 for the impact of the number of retrieved procedural memories ($k$). 
We evaluated the success rate across four configurations: the $\pi_{0.5}$ baseline with no fused adapter, and {VLA-Pro} with $k \in \{1, 2, 3\}$.
As illustrated in Fig.~\ref{fig:topk_analysis}, {VLA-Pro ($k$=2)} achieves the optimal average success rate of 20.9\%, representing a 51.4\% improvement over the baseline (13.8\%). 
The results indicate that effective cross-task transfer often requires integrating knowledge from a proper number of related experiences rather than relying on a single source.


\section{Conclusion}
\label{sec:conclusion}

In this work, we introduced VLA-Pro, a procedural memory transfer framework for improving the cross-task generalization of Vision-Language-Action models. By representing each execution stage with structured procedural states and storing task-specific experience as lightweight LoRA memories, VLA-Pro enables action-aware retrieval and dynamic parameter fusion during inference. Experiments on RoboTwin, RLBench, and real-world manipulation tasks demonstrate that the proposed framework consistently improves unseen-task performance across different VLA backbones, showing the effectiveness of procedural memory as a mechanism for transferring manipulation experience.

Despite these promising results, VLA-Pro still has several limitations. Its performance depends on the quality of procedural state extraction and memory retrieval, which may be affected by ambiguous observations, incomplete interaction history, or inaccurate multimodal reasoning. In addition, the current memory bank is constructed from a limited set of seen tasks, leaving open the question of how to scale procedural memories to larger and more diverse robotic experience libraries. Future work will explore more robust state extraction, continual memory expansion, and more efficient fusion strategies for long-horizon and highly dynamic real-world manipulation tasks.

\clearpage

\bibliographystyle{plainnat}
\bibliography{main}





\newpage

\appendix

\section{Procedural Memory Storage and Retrieval}
\label{app:storage_retrieval}

\subsection{Prompt for Procedural State Extraction}
\label{app:prompt}

We use the following prompt to extract the current procedural state from the visual observation and language instruction. The output is constrained to a structured JSON format following existing approaches~\cite {ye2025learning}, which is then used for procedural memory retrieval.

\begin{tcolorbox}[
    enhanced,
    breakable,
    colback=blue!3,
    colframe=blue!25,
    boxrule=0.5pt,
    arc=2pt,
    left=6pt,
    right=6pt,
    top=6pt,
    bottom=6pt,
    title={System Prompt},
    fonttitle=\bfseries,
    coltitle=black,
    colbacktitle=blue!8
]
\begin{Verbatim}[breaklines=true,breakanywhere=true,fontsize=\small]
Role: Embodied AI subtask target-point predictor with vision-language input.

Task: Given one image and one instruction, output ONE JSON describing the CURRENT subtask target point only. Do not output any extra text.

Output must be a single strictly valid JSON object with exactly this schema:
{
  "subtask": "sentence",
  "action": "word",
  "entity_shape": "word",
  "ee_orientation": "word",
  "target_point": "word"
}

Allowed Definitions:
- subtask: A concise verb-noun phrase describing the current subtask.

Allowed Enum Values:
- action: [options]
   - (constraints)

- entity_shape: [options]
   - (constraints)

- ee_orientation: [options]
   - (constraints)

- target_point: [options]
   - (constraints)

Hard constraints:
1) Output JSON only. No markdown, no comments, no trailing commas.
2) Do not add/remove keys. Keep exactly the keys shown above.
3) Use ONLY one of the allowed enum values.
\end{Verbatim}
\end{tcolorbox}

\subsection{Procedural State Schema and Matching Weights}
\label{app:state_schema_matching}

This section summarizes the procedural state schema for RoboTwin tasks and the matching weights used in Action-Aware Procedural Matching. The free-form \texttt{subtask} field is only used for readability and debugging, and is excluded from similarity computation to avoid semantic interference from surface-level language descriptions.

\begin{tcolorbox}[
    enhanced,
    breakable,
    colback=blue!3,
    colframe=blue!25,
    boxrule=0.5pt,
    arc=2pt,
    left=6pt,
    right=6pt,
    top=6pt,
    bottom=6pt,
    title={Procedural State Schema},
    fonttitle=\bfseries\small,
    coltitle=black,
    colbacktitle=blue!8
]
\small
\begin{tabular}{p{0.08\linewidth} p{0.20\linewidth} p{0.62\linewidth}}
\toprule
\textbf{Symbol} & \textbf{Field} & \textbf{Candidate Values / Description} \\
\midrule
-- & \texttt{subtask} 
& A concise natural-language phrase for readability only; not used for similarity computation. \\

$a$ & \texttt{action} 
& \texttt{pick}, \texttt{place}, \texttt{press}, \texttt{push}, \texttt{drag}. \\

$o$ & \texttt{entity\_shape} 
& \texttt{open\_container}, \texttt{cuboid}, \texttt{spherical}, \texttt{handle}, \texttt{lying\_cylindrical}, \texttt{upright\_cylindrical}, \texttt{other}. \\

$e$ & \texttt{ee\_orientation} 
& \texttt{vertical}, \texttt{horizontal}. \\

$p$ & \texttt{target\_point} 
& \texttt{front}, \texttt{back}, \texttt{left}, \texttt{right}, \texttt{center}, \texttt{midpoint}, \texttt{end}, \texttt{top}, \texttt{rim}. \\
\bottomrule
\end{tabular}
\end{tcolorbox}

\par\medskip
\noindent
\begin{minipage}{\linewidth}
For example, the first action chunk of the RoboTwin
\texttt{place\_upright\_stand} task has the following procedural state:
\par\smallskip

{\small\ttfamily\raggedright
\{\\
\quad "subtask": "pick the bottle",\\
\quad "action": "pick",\\
\quad "entity\_shape": "upright\_cylindrical",\\
\quad "ee\_orientation": "horizontal",\\
\quad "target\_point": "center"\\
\}
\par}
\end{minipage}
\par\medskip

For matching, each selected field-value pair is encoded into a text embedding, and the action-dependent weights emphasize the procedural attribute most relevant to the current action.

\begin{tcolorbox}[
    enhanced,
    breakable,
    colback=green!3,
    colframe=green!25,
    boxrule=0.5pt,
    arc=2pt,
    left=6pt,
    right=6pt,
    top=6pt,
    bottom=6pt,
    title={Action-Aware Matching Weights},
    fonttitle=\bfseries\small,
    coltitle=black,
    colbacktitle=green!8
]
\small
\begin{tabular}{p{0.22\linewidth} p{0.26\linewidth} p{0.44\linewidth}}
\toprule
\textbf{Condition} & \textbf{Emphasized Field} & \textbf{Weight Setting} \\
\midrule
Default & action consistency & $w_a=0.35$, $w_o=w_e=w_p=0.15$ \\
$a=\texttt{pick}$ & object geometry & $w_o=0.35$ \\
$a=\texttt{place}$ & target interaction point & $w_p=0.35$ \\
$a=\texttt{push}$ & end-effector orientation & $w_e=0.35$ \\
\bottomrule
\end{tabular}
\end{tcolorbox}

\section{Task Suite and Data Construction}
\label{app:task_suite}

\subsection{RoboTwin Task Suite}
\label{app:robotwin_tasks}

\begin{figure*}[htbp] 
    \centering
    \includegraphics[width=\textwidth]{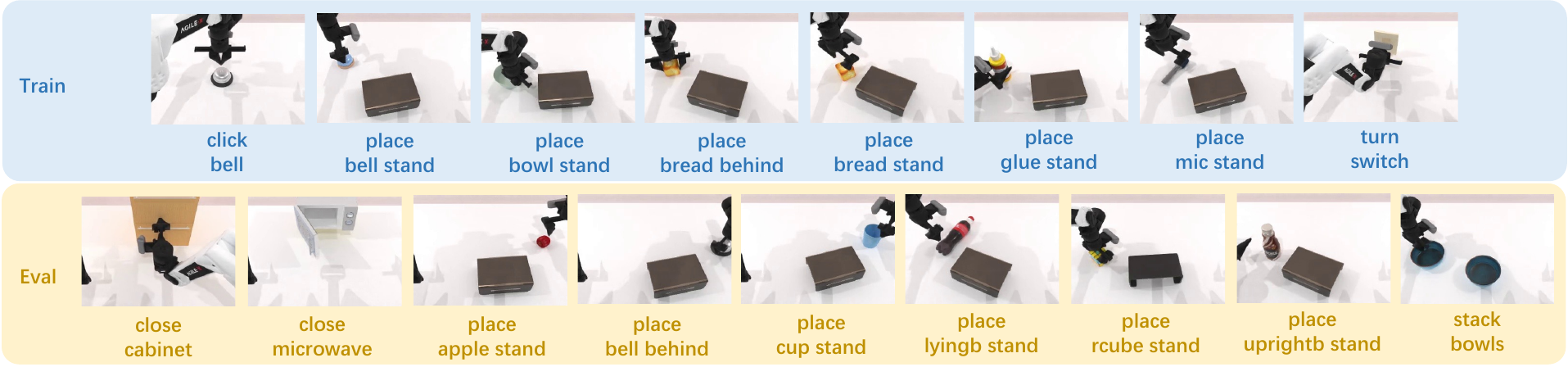}
    \caption{Visualization of RoboTwin dataset construction. This figure illustrates the 8 training tasks and the corresponding 9 test tasks for cross-task generalization evaluation.}
    \label{fig:robotwin_data}
\end{figure*}

\par\medskip
\noindent
    \begin{minipage}[t]{0.70\linewidth}
        \vspace{0pt}
        \noindent We build our customized RoboTwin task suite by modifying the original RoboTwin environment. Specifically, we retain 2 original training tasks and construct 6 additional training tasks and 9 held-out test tasks. The full task list and reference instructions are provided in Tables~\ref{tab:robotwin_training_tasks} and~\ref{tab:robotwin_test_tasks}.

        \vspace{0.5em}
        The listed instructions are reference forms only; following RoboTwin, each task has 50 language variants. The test tasks share related objects, actions, or spatial relations with the training tasks, while remaining unseen during training, enabling evaluation of cross-task procedural transfer. For example, \texttt{place\_bell\_behind} tests whether the model can avoid the dominant ``place-on-stand'' memory and retrieve the correct spatial-relation procedure. The pair of \texttt{place\_uprightb\_stand} and \texttt{place\_lyingb\_stand} further evaluates whether the model can distinguish object states despite semantic similarity in task names. The overall task-suite construction is visualized in Fig.~\ref{fig:robotwin_data}.
    \end{minipage}
    \hfill
    \begin{minipage}[t]{0.26\linewidth}
        \vspace{0pt}
        \centering
        \includegraphics[width=\linewidth]{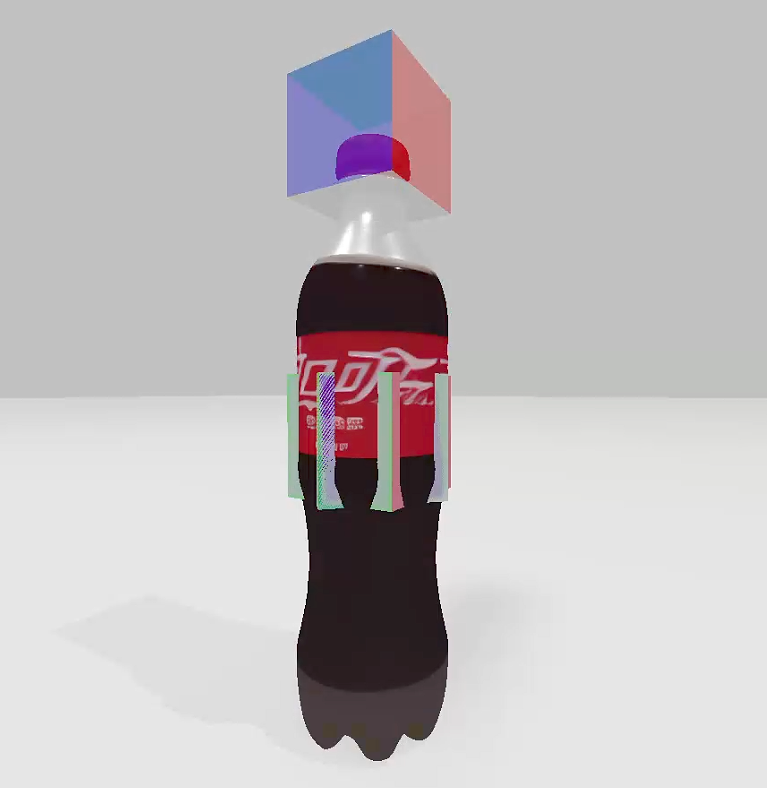}
        \captionof{figure}{Modified grasp-point configuration for
        constructing procedurally related RoboTwin tasks.}
        \label{fig:robotwin_bottle_model}
    \end{minipage}
\par\medskip

\begin{table}[htbp]
\centering
\small
\setlength{\tabcolsep}{6pt}
\renewcommand{\arraystretch}{1.12}
\caption{Training tasks and language instructions in our customized RoboTwin task suite.}
\label{tab:robotwin_training_tasks}
\begin{tabular}{p{0.34\linewidth} p{0.58\linewidth}}
\toprule
\textbf{Task Name} & \textbf{Reference Language Instruction} \\
\midrule
\texttt{place\_bell\_stand} & Place the bell on the stand. \\
\texttt{place\_bowl\_stand} & Grab the bowl and place it on the display stand. \\
\texttt{place\_bread\_behind} & Place the bread behind the display stand. \\
\texttt{place\_bread\_stand} & Pick up the bread and put it on the display stand. \\
\texttt{place\_glue\_stand} & Place the glue on the display stand. \\
\texttt{place\_mic\_stand} & Place the microphone on the display stand. \\
\texttt{turn\_switch} & Press the switch to activate it. \\
\texttt{click\_bell} & Press the top center of the bell. \\
\bottomrule
\end{tabular}
\end{table}


\begin{table}[htbp]
\centering
\small
\setlength{\tabcolsep}{4pt}
\renewcommand{\arraystretch}{1.10}
\caption{Testing tasks and language instructions in our customized RoboTwin task suite.}
\label{tab:robotwin_test_tasks}
\resizebox{\linewidth}{!}{%
\begin{tabular}{@{}lll@{}}
\toprule
\textbf{Task Name} & \textbf{Reference Language Instruction} & \textbf{Main Target Source Memory} \\
\midrule
\texttt{place\_apple\_stand} 
& Pick up the apple and place it on the display stand. 
& \texttt{place\_bell\_stand} \\

\texttt{place\_bell\_behind} 
& Place the bell behind the display stand. 
& \texttt{place\_bread\_behind} \\

\texttt{place\_cup\_stand} 
& Place the cup on the display stand. 
& \texttt{place\_bowl\_stand} \\

\texttt{place\_lyingb\_stand} 
& Place the lying bottle on the display stand. 
& \texttt{place\_mic\_stand} \\

\texttt{place\_rcube\_stand} 
& Place the Rubik's cube on the display stand. 
& \texttt{place\_bread\_stand} \\

\texttt{place\_uprightb\_stand} 
& Place the upright bottle on the display stand. 
& \texttt{place\_glue\_stand} \\

\texttt{stack\_bowls} 
& Lift one bowl and place it on the other bowl. 
& \texttt{place\_bowl\_stand} \\

\texttt{close\_cabinet} 
& Close the cabinet. 
& \texttt{turn\_switch} \\

\texttt{close\_microwave} 
& Close the microwave. 
& \texttt{turn\_switch} \\
\bottomrule
\end{tabular}%
}
\end{table}

\subsection{RLBench Task Suite}
Our RLBench experiments follow the X-ICM~\citep{zhouexploring} task split, which contains 18 training tasks and 23 held-out test tasks for cross-task generalization evaluation. From the 18 training tasks, we select 8 foundational tasks as source memories, since they cover basic procedural elements. The selected source-memory tasks are:
\texttt{close\_jar}, \texttt{open\_drawer}, \texttt{place\_wine\_at\_rack\_location}, \texttt{push\_buttons}, \texttt{put\_groceries\_in\_cupboard}, \texttt{put\_item\_in\_drawer}, \texttt{reach\_and\_drag}, and \texttt{stack\_blocks}. These source memories are used to evaluate whether VLA-Pro can retrieve and transfer relevant procedural memory to unseen RLBench tasks.

\subsection{Real-World Task Suite}
\label{app:realworld_tasks}

This section provides additional details about the real-world task suite. We design 6 training tasks and 6 corresponding held-out test tasks to evaluate whether the model can reuse relevant procedural memories while avoiding interference from incorrect memories. The training and test tasks are listed in Tables~\ref{tab:realworld_training_tasks} and~\ref{tab:realworld_test_tasks}, respectively.

\begin{figure*}[hptb]
    \centering
    \includegraphics[width=\textwidth]{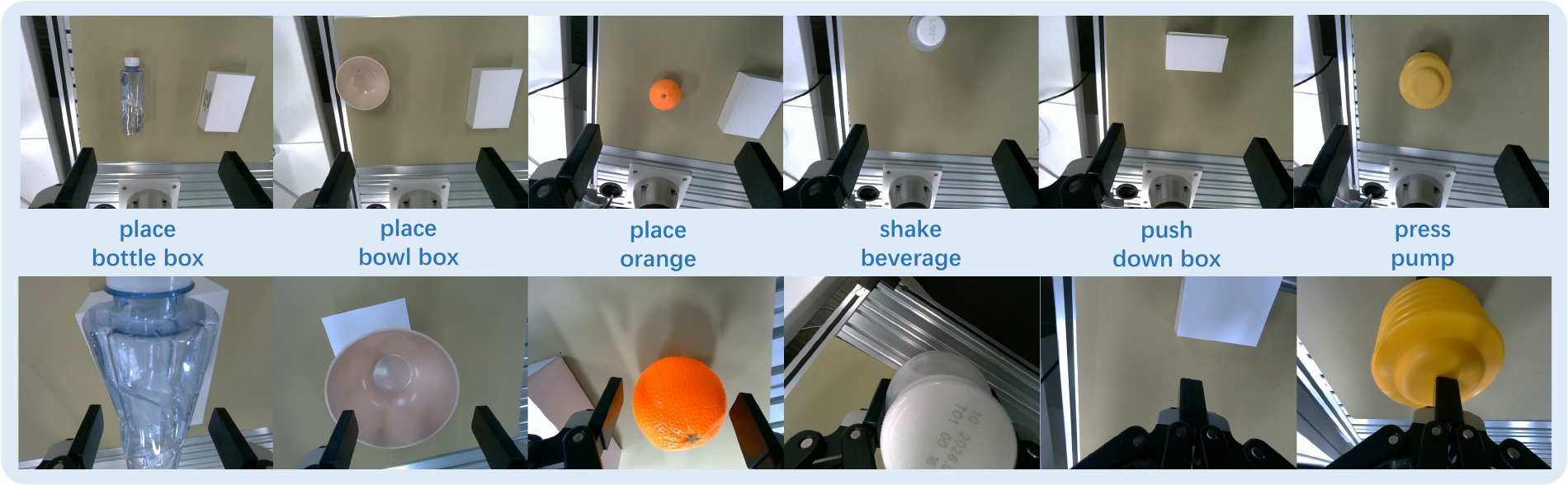}
    \caption{Examples of initial and final wrist-camera images for the real-world training tasks. As the visual observation changes during task execution, the model infers the current execution stage accordingly and retrieves the relevant procedural memory.}
    \label{fig:realworld_training_stage_examples}
\end{figure*}

\begin{table}[htbp]
\centering
\small
\setlength{\tabcolsep}{6pt}
\renewcommand{\arraystretch}{1.12}
\caption{Real-world training tasks and corresponding reference language instructions.}
\label{tab:realworld_training_tasks}
\begin{tabular}{p{0.34\linewidth} p{0.58\linewidth}}
\toprule
\textbf{Task Name} & \textbf{Reference Language Instruction} \\
\midrule
\texttt{place\_bottle\_box} & Pick up the bottle and place it on the box. \\
\texttt{place\_bowl\_box} & Pick up the bowl and place it on the white box. \\
\texttt{place\_orange\_box\_right} & Pick up the orange and place it on the right side of the box. \\
\texttt{shake\_beverage} & Pick up the beverage bottle and shake it. \\
\texttt{press\_pump} & Press the pump. \\
\texttt{push\_down\_box} & Push down the box. \\
\bottomrule
\end{tabular}
\end{table}


\begin{table}[htbp]
\centering
\small
\setlength{\tabcolsep}{4pt}
\renewcommand{\arraystretch}{1.10}
\caption{Real-world testing tasks and corresponding reference language instructions.}
\label{tab:realworld_test_tasks}
\resizebox{\linewidth}{!}{%
\begin{tabular}{@{}lll@{}}
\toprule
\textbf{Task Name} & \textbf{Reference Language Instruction} & \textbf{Main Target Source Memory} \\
\midrule
\texttt{place\_microphone\_box} 
& Pick up the microphone and place it on the box. 
& \texttt{place\_bottle\_box} \\

\texttt{place\_cup\_box} 
& Pick up the cup and place it on the box. 
& \texttt{place\_bowl\_box} \\

\texttt{place\_bottle\_box\_right} 
& Grab the bottle and place it on the right of the box. 
& \texttt{place\_orange\_box\_right} \\

\texttt{shake\_chemical} 
& Pick up the chemical bottle and shake it. 
& \texttt{shake\_beverage} \\

\texttt{tap\_chips\_can} 
& Tap the top of the chips can. 
& \texttt{press\_pump} \\

\texttt{flick\_bottle} 
& Flick the plastic bottle. 
& \texttt{push\_down\_box} \\
\bottomrule
\end{tabular}%
}
\end{table}

\section{Training Hyperparameters and Implementation Details}
\label{app:implementation_details}

\noindent\textbf{RoboTwin.}
For RoboTwin experiments, VLA-Pro is implemented on three backbones: X-VLA, RDT-1B, and $\pi_{0.5}$. For X-VLA, training starts from \texttt{2toINF/X-VLA-SoftFold}. LoRA adapters are trained with learning rate $1\times10^{-5}$, batch size 4, gradient accumulation steps 8, and effective batch size 32. The Adam optimizer uses $\beta=(0.9,0.95)$, weight decay 0.0, and maximum gradient norm 1.0. The LoRA rank and alpha are set to 128 and 256, respectively. The Base LoRA baseline is trained for 10,000 steps, and each VLA-Pro task-specific adapter is further trained for 2,000 steps on 4 NVIDIA RTX 4090 GPUs with 24GB memory. LoRA is applied to the backbone linear layers, including the merged attention \texttt{qkv} projection.

For RDT-1B, the official pretrained RDT-1B checkpoint is used. The model is trained with AdamW using a constant learning rate of $1\times10^{-4}$ and batch size 20. The Base LoRA baseline is trained for 5,000 steps, and each VLA-Pro task-specific adapter is trained for 1,000 steps. The LoRA rank and alpha are set to 128 and 32, respectively. Training is conducted on 6 NVIDIA RTX 4080 GPUs with 32GB memory. LoRA is applied to the linear modules \texttt{qkv}, \texttt{proj}, \texttt{q}, \texttt{kv}, \texttt{fc1}, and \texttt{fc2}, covering both attention projections and feed-forward layers.

For $\pi_{0.5}$, the official \texttt{pi05\_base} checkpoint is used. Training adopts AdamW with a cosine learning-rate schedule using 1,000 warmup steps and a peak learning rate of $2.5\times10^{-5}$. The LoRA rank and alpha are both set to 32. The Base LoRA baseline is trained for 5,000 steps, and each VLA-Pro task-specific adapter is trained for 3,000 steps on 6 NVIDIA RTX 4080 GPUs with 32GB memory. For all $\pi_{0.5}$ experiments, LoRA is applied to the attention and feed-forward layers of the Gemma backbone and action-expert, covering both attention projections and MLP projections.
\begin{figure*}[htbp]
    \centering
    \includegraphics[width=\textwidth]{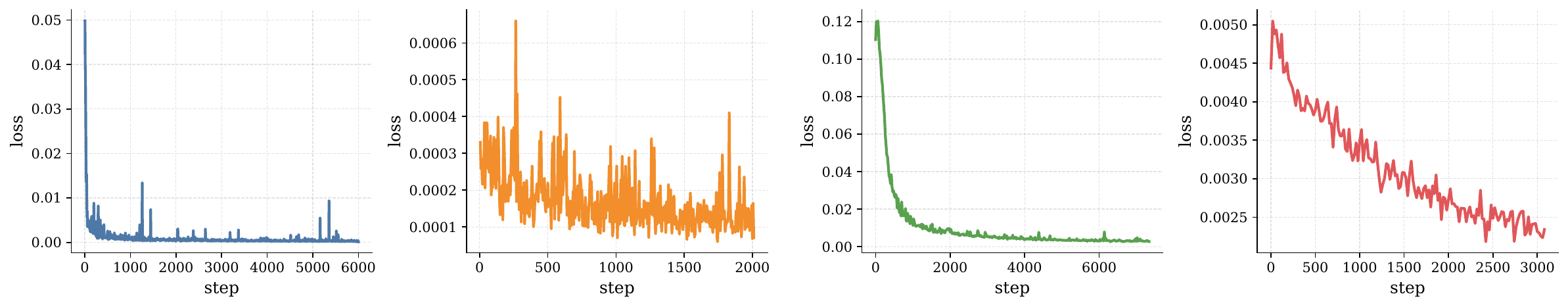}
    \caption{Representative training loss curves in the RoboTwin experiments. From left to right, the plots show the loss curve of the RDT baseline, the continued VLA-Pro training with RDT as the backbone, the $\pi_{0.5}$ baseline, and the continued VLA-Pro training with $\pi_{0.5}$ as the backbone.}
    \label{fig:robotwin_training_loss}
\end{figure*}

\noindent\textbf{RLBench.}
In RLBench experiments, RDT uses the official RDT-1B pretrained model. For AtomicVLA, AdamW is used with a cosine learning-rate schedule using 10,000 warmup steps and a peak learning rate of $5\times10^{-5}$. The batch size is 32, the number of experts is set to 5, and the model is trained for 5,000 steps on 2 NVIDIA H100 GPUs with 94GB memory.

For the $\pi_{0.5}$ Base LoRA baseline, AdamW is used with the same cosine learning-rate schedule as in the RoboTwin $\pi_{0.5}$ experiments, with 1,000 warmup steps and a peak learning rate of $2.5\times10^{-5}$. The batch size is 64, and the baseline is trained for 50,000 steps. For VLA-Pro, the same optimizer and learning-rate schedule are used. Starting from the Base LoRA baseline, each task-specific adapter is further trained for 6,000 steps. Training is conducted on 4 NVIDIA RTX 5880 GPUs with 48GB memory.

\noindent\textbf{Real-world Experiments.}
For real-world experiments, $\pi_{0.5}$ is used as the backbone and the Base LoRA serves as the baseline. Training uses AdamW with a cosine learning-rate schedule using 1,000 warmup steps and a peak learning rate of $2.5\times10^{-5}$. The LoRA rank and alpha are both set to 32, and the batch size is 64. The Base LoRA baseline is trained for 10,000 steps, and VLA-Pro further trains each task-specific adapter for 3,000 steps. Training is conducted on 2 NVIDIA A100 GPUs with 80GB memory.

\section{Limitations and Future Work}

Although VLA-Pro demonstrates consistent improvements across simulation and real-world manipulation tasks, it still has several limitations. First, the effectiveness of procedural memory transfer depends on the quality of procedural state extraction and memory retrieval. When the visual observation is ambiguous, the interaction history is incomplete, or the extracted procedural state is inaccurate, VLA-Pro may retrieve less relevant memories and inject suboptimal adapters. Second, the current memory bank is constructed from a limited set of seen tasks, which restricts the diversity of reusable procedural experiences. More importantly, although VLA-Pro is motivated by human procedural memory, it still cannot use memory as flexibly as humans. Humans can often abstract, recombine, and adapt past experiences across substantially different contexts, whereas VLA-Pro mainly reuses task-specific LoRA memories through similarity-based retrieval and weighted fusion. This makes the current framework effective for related cross-task transfer, but less capable of handling highly novel tasks that require deeper compositional reasoning or creative memory adaptation.

Future work will explore more flexible and scalable forms of procedural memory. One direction is to build larger and continuously expandable memory banks from more diverse robotic experiences, allowing the model to retrieve from a broader range of manipulation patterns. Another direction is to improve the memory representation itself, for example by introducing hierarchical or compositional procedural states that can better capture reusable subskills, object affordances, and action constraints. In addition, future systems could incorporate uncertainty-aware retrieval and adaptive memory fusion, so that the model can decide when to rely on retrieved memories, when to combine multiple memories, and when to fall back to the base policy. These improvements may enable VLA models to move closer to human-like memory utilization, where past experiences are not only retrieved, but also flexibly reorganized and adapted to novel real-world situations.


\end{document}